%% file: main.tex
\pdfoutput=1

\documentclass[11pt]{article}

\usepackage{acl}

\usepackage{times}
\usepackage{latexsym}
\usepackage{multirow}
\usepackage{graphicx}
\usepackage{booktabs}
\usepackage{algorithm}
\usepackage{algorithmic}
\usepackage[T1]{fontenc}

\usepackage[utf8]{inputenc}

\usepackage{microtype}

\usepackage{caption}
\usepackage{subcaption}

\input{math_commands}

\usepackage{soul, color, xcolor}
\usepackage{mathptmx}
\newcommand\norm[1]{\lVert#1\rVert}

\newcommand{\upp}[1]{\textcolor{black}{#1}} 
\newcommand{\uu}[1]{\textcolor{black}{#1}} 
\newcommand{\uuu}[1]{\textcolor{black}{#1}} 

\newcommand{\methodname}[1]{iACE} 
\newcommand{\problemname}[1]{ICLU}

\title{Imagination-Augmented Natural Language Understanding}

\author{
Yujie Lu$^1$, Wanrong Zhu$^1$, Xin Eric Wang$^2$, Miguel Eckstein$^1$, William Yang Wang$^1$\\
$^1$University of California, Santa Barbara, CA, USA\\ 
\texttt{\{yujielu,wanrongzhu,migueleckstein,wangwilliamyang\}@ucsb.edu}\\ 
$^2$University of California, Santa Cruz, CA, USA\\
\texttt{xwang366@ucsc.edu}\\
}

\begin{document}
\maketitle
\begin{abstract}
Human brains integrate linguistic and perceptual information simultaneously to understand natural language, and hold the critical ability to render imaginations.
Such abilities enable us to construct new abstract concepts or concrete objects, and are essential in involving practical knowledge to solve problems in low-resource scenarios.  
However, most existing methods for Natural Language Understanding (NLU) are mainly focused on textual signals. They do not simulate human visual imagination ability, which hinders models from inferring and learning efficiently from limited data samples.
Therefore, we introduce an \textbf{I}magination-\textbf{A}ugmented \textbf{C}ross-modal \textbf{E}ncoder (\methodname~) to solve natural language understanding tasks from a novel learning perspective---imagination-augmented cross-modal understanding.
\methodname~ enables visual imagination with external knowledge transferred from the powerful generative and pre-trained vision-and-language models.
Extensive experiments on GLUE \cite{Wang2018GLUEAM} and SWAG \cite{Zellers2018SWAGAL} show that \methodname~ achieves consistent improvement over visually-supervised pre-trained models.
More importantly, results in extreme and normal few-shot settings validate the effectiveness of \methodname~ in low-resource natural language understanding circumstances.\footnote{Source code and pre-trained models are publicly available at \href{https://github.com/YujieLu10/IACE-NLU}{https://github.com/YujieLu10/IACE-NLU}}

\end{abstract}

\input{sections/1-introduction}
\input{sections/2-related}

\input{sections/3-method}
\input{sections/4-experiments}

\input{sections/5-analysis}
\input{sections/6-conclusion}


\bibliography{anthology,custom}
\bibliographystyle{acl_natbib}

\end{document}

%% file: math_commands.tex
\usepackage{amsmath,amsfonts,bm}

\def\eqref#1{equation~\ref{#1}}

\def\1{\bm{1}}

\def\vi{{\bm{i}}}

\def\vt{{\bm{t}}}

\def\vv{{\bm{v}}}

\DeclareMathAlphabet{\mathsfit}{\encodingdefault}{\sfdefault}{m}{sl}
\SetMathAlphabet{\mathsfit}{bold}{\encodingdefault}{\sfdefault}{bx}{n}

\def\gI{{\mathcal{I}}}

\def\gX{{\mathcal{X}}}



%% file: sections/1-introduction.tex
\section{Introduction}
Cognitive neuroscience studies reveal neural activation in vision-related brain areas when reading text~\citep{Just2004ImageryIS} and show a tight relationship between brain areas processing linguistic and visual semantic information~\citep{popham2021cortex}. In addition, visual imagery improves comprehension during human language processing~\citep{Sadoski1994ADC}. \uuu{Such imagination empowers human brains with generalization capability to solve problems with limited supervision or data samples.}

However, the field of Natural language Understanding has mainly been focused on building machines based solely on language, ignoring the inherently grounded imagination from the external visual world.
These studies either learn text-only representations from language corpora~\citep{Devlin2019BERTPO, Liu2019RoBERTaAR, Lan2020ALBERTAL} or implicitly involve retrieved visual supervision in pre-trained language models~\citep{Tan2020VokenizationIL}.
Thus, their approaches appear limited in transferring the connection between language understanding and visual imagination to downstream tasks, which are essential to solving low-resource circumstances.
In addition, these methods are limited to text-only augmentations, whereas visual imaginations leverage cross-modal augmentations to deal with low-resource situations.

\begin{figure}[t]
    \centering
    \includegraphics[width=\linewidth]{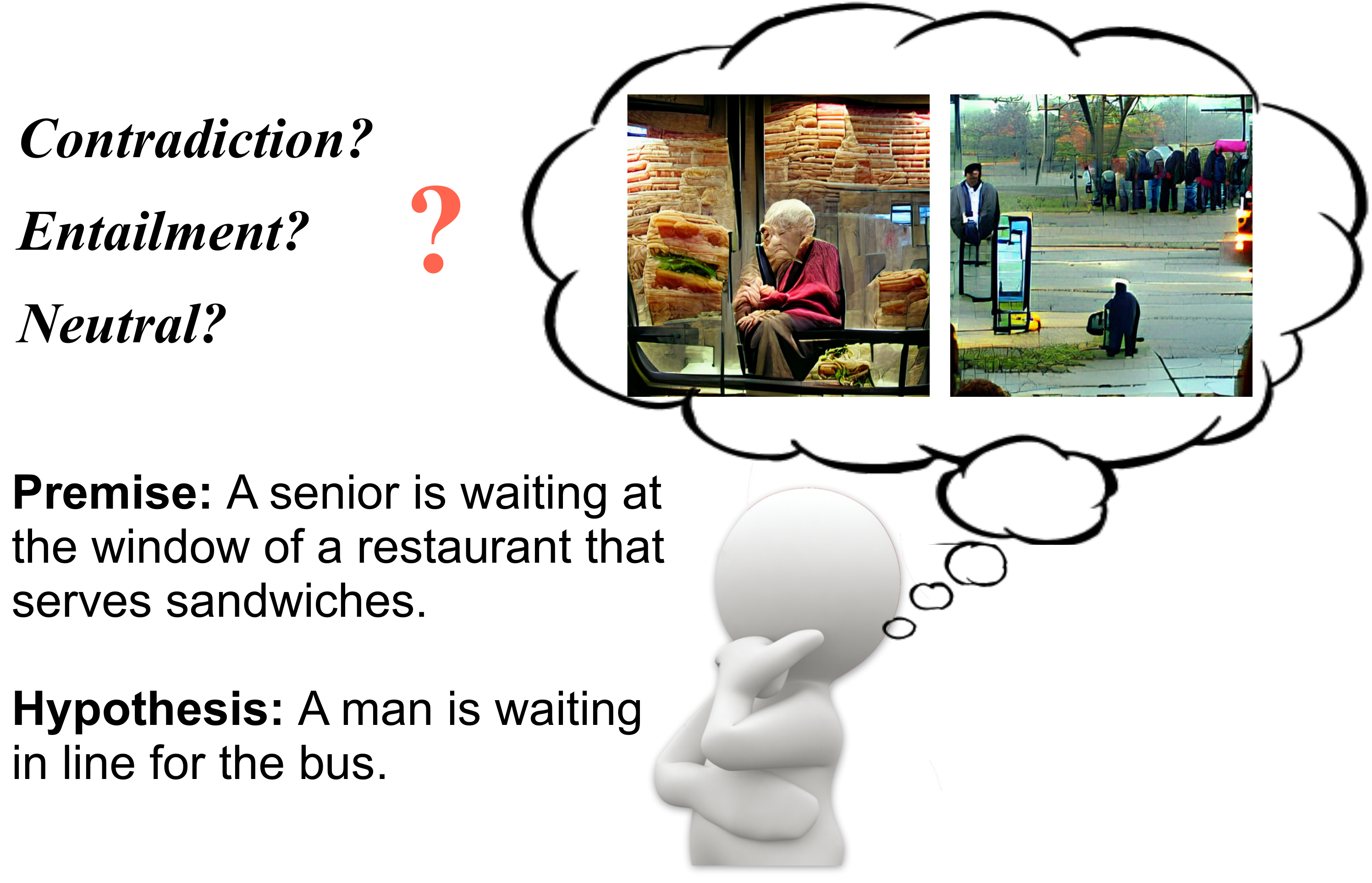}
    \caption{
    \upp{Rendering visual imagination is an intuitive way to activate perception for linguistic understanding, e.g. natural language inference.}
    }
    \label{fig:humanlike}
\end{figure}

Human brains are multi-modal, integrating linguistic and perceptual information simultaneously.
Intuitively, the \uuu{machines} could achieve a higher-level understanding of natural language and better learning transference by imitating the procedure of human imagination behavior.

\begin{figure}[t]
    \centering
    \includegraphics[width=0.9\linewidth]{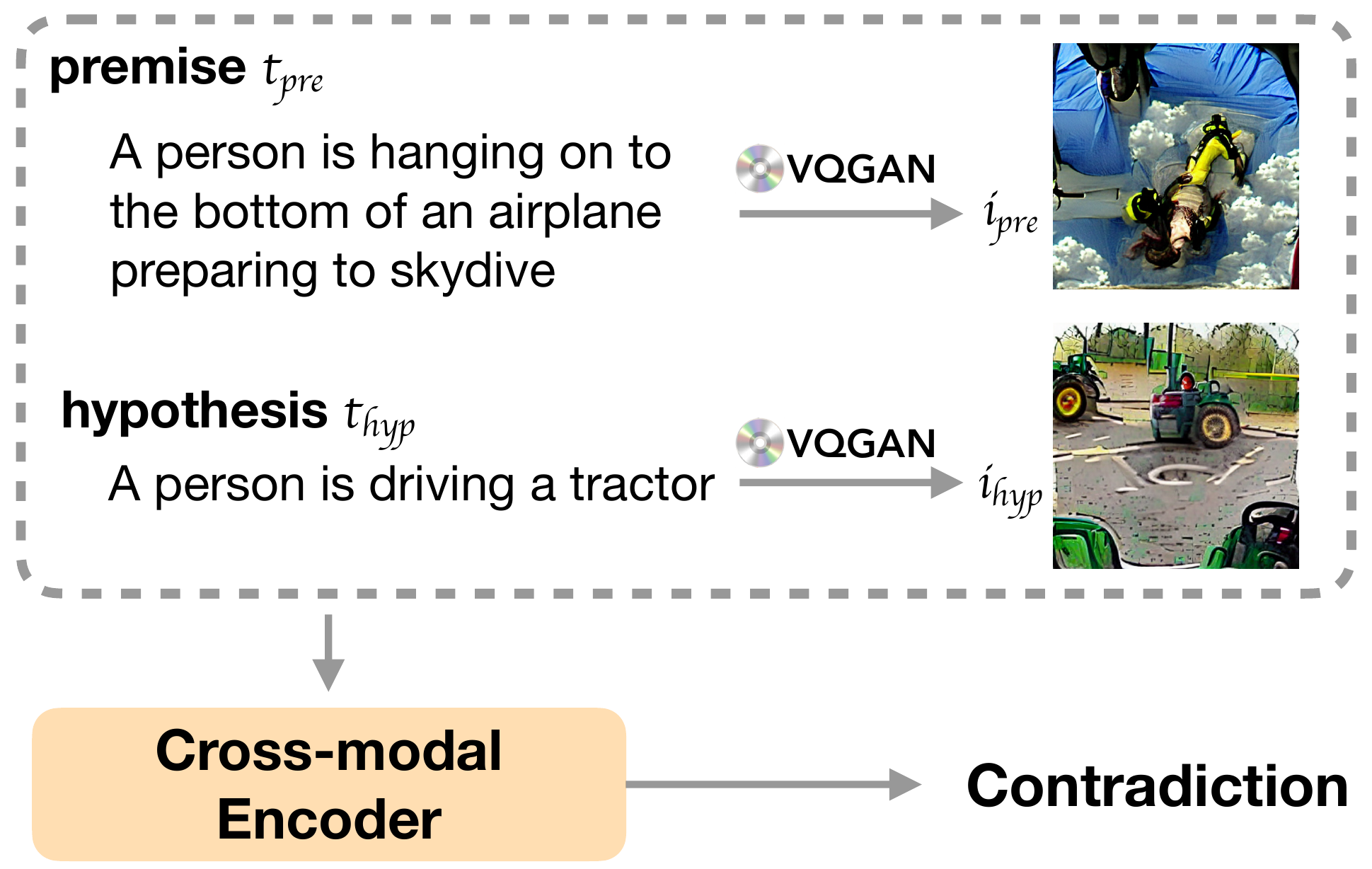}
    \caption{
     A detailed view of our \methodname~ framework fine-tunes on natural language inference task.
    }
    \label{fig:task}
\end{figure}

\upp{Inspired by this, we propose to understand language with the integration of linguistic and perceptual information via introducing imagination supervision into text-only NLU tasks.
To imitate the imagination-augmented understanding process as shown in Figure~\ref{fig:humanlike} with text-only data, we devise a procedure with two steps: 1) pre-train a visually-supervised Transformer over paired text and images retrieved from large-scale language corpus and image set, and 2) construct the imagination with a generative model and fine-tune on downstream NLU datasets by learning the paired imagination and natural language in a cross-modal embedding.
We show a detailed description of the cross-modal imagination process for a specific Natural Language Inference task in Figure~\ref{fig:task}.
In this way, we utilize machine imagination to improve the performance of natural language understanding.}

We adopt the few-shot learning setting to study the potential of using less human effort of annotation for our proposed \methodname~ to learn the natural language with the help of imagination.
Large margin performance gain in both extreme and normal few-shot settings demonstrate the effectiveness of \methodname~ in solving problems with limited data samples.
In the full data setting of GLUE~\cite{Wang2018GLUEAM} and SWAG~\cite{Zellers2018SWAGAL}, we observe the consistent performance gain of our proposed \methodname~ over the visually-supervised approach (e.g., VOKEN~\cite{Tan2020VokenizationIL}) upon four language base models (e.g., BERT, RoBERTa).

In summary, the main contributions of our work are as follows:
\begin{itemize}
    \item
    We propose to solve the text-only learning problem in natural language understanding tasks from a novel learning perspective: imagination-augmented cross-modal language understanding.
    \item To address the problem mentioned above, we devise \methodname~ to generate imaginations in a cross-modal representation space to guide the fine-tuning of the visually supervised language models.
    \item \upp{Experimental results in the few-shot setting validate the consistent superiority of \methodname~ over baselines in tackling the low-resource situation. In full settings, \methodname~ maintains the improvement in GLUE and SWAG.}
\end{itemize}

%% file: sections/2-related.tex
\section{Related Work}
\paragraph{Visually-aided Language Learning}
Previous research attempt to introduce visual information to improve language learning on various Natural Language Processing (NLP) scenarios, including but not limited to machine translation~\citep{Grubinger2006TheIT,Elliott2016Multi30KME}, information retrieval~\citep{Funaki2015ImageMediatedLF,Gu2018LookIA}, semantic parsing~\citep{Christie2016ResolvingLA,Shi2019VisuallyGN}, natural language inference~\citep{Xie2019VisualEA}, bilingual lexicon learning~\citep{Kiela2015VisualBL,Vulic2016MultiModalRF}, natural language generation evaluation~\citep{Zhu2021ImaginEAI}, spatial commonsense reasoning~\citep{da2022SpatialCommonsense} and language representation learning~\citep{Lazaridou2015CombiningLA,Collell2017ImaginedVR,Kiela2018LearningVG,Zablocki2019IncorporatingVS,Lu2019ViLBERTPT,Li2019VisualBERTAS,Sun2019VideoBERTAJ,Luo2020UniViLMAU,Chen2020UNITERUI,Li2020UnicoderVLAU,Tan2020VokenizationIL,Radford2021LearningTV}.
While most of these studies acquire visual information through retrieval from the web or large-scale image sets, a recent line of studies attempt to generate visual supervision from scratch. The visual information can either be provided in the form of representation~\citep{Collell2017ImaginedVR,Long2021GenerativeIE} or concrete images~\citep{Gu2018LookIA,Zhu2021ImaginEAI}.
Though previous studies generate machine imagination, they only tackle specific tasks, such as machine translation~\citep{Long2021GenerativeIE} or information retrieval~\citep{Gu2018LookIA}.
To the best of our knowledge, we are the first to utilize machine abstract imagination from large pretrained vision and language models to improve general NLU tasks.
\uuu{Recently, VOKEN~\cite{Tan2020VokenizationIL} incorporate retrieved token-level visual information into existing transformer models and achieve consistent improvement.
\methodname~ is different from this work for two aspects: 1) we explicitly encode visual imagination during fine-tuning. 2) we propose a novel model to borrow knowledge from imagination in both training and inference.}

\paragraph{Few-shot Natural Language \upp{Understanding}}
Natural Language Understanding (NLU) is a subfield in NLP that involves a broad range of tasks such as question answering, sentiment analysis, and textual entailment. Researchers have collected specific language corpus~\citep{Wang2018GLUEAM,Zellers2018SWAGAL,Xu2020CLUEAC} to train the machines on NLU learning. However, the general language understanding problem remains a challenge.
Few-shot learning is a learning paradigm that aims to predict the correct class of instances with a relatively small amount of labeled training examples~\citep{Fink2004ObjectCF,FeiFei2006OneshotLO}. It has been receiving increasing attention for its potential in reducing data collection effort and computational costs and extending to rare cases.
To deal with data-scarcity in NLU problems, previous research introduces external knowledge~\citep{Sui2021KnowledgeGM}, utilizes meta-learning~\citep{Geng2019InductionNF,Bansal2020LearningTF,Han2021MetaLearningAD} and adopts data augmentation to generate labeled utterances for few-shot classes~\citep{Murty2021DReCaAG,Wei2021FewShotTC}.
Recent studies~\citep{Radford2019LanguageMA,Brown2020LanguageMA} have shown that large-scale pre-trained language models are able to perform NLU tasks in a few-shot learning manner. 
The pre-trained multimodal models also display similar few-shot learning ability~\citep{Tsimpoukelli2021MultimodalFL}.
Unlike previous studies on pre-trained multimodal Transformers that target solving multimodal tasks, our study introduces imagination from the visual world into language models and aims to improve NLU.

%% file: sections/3-method.tex
\section{Our Approach}

\begin{figure*}[t]
    \centering
    \includegraphics[width=0.88\linewidth]{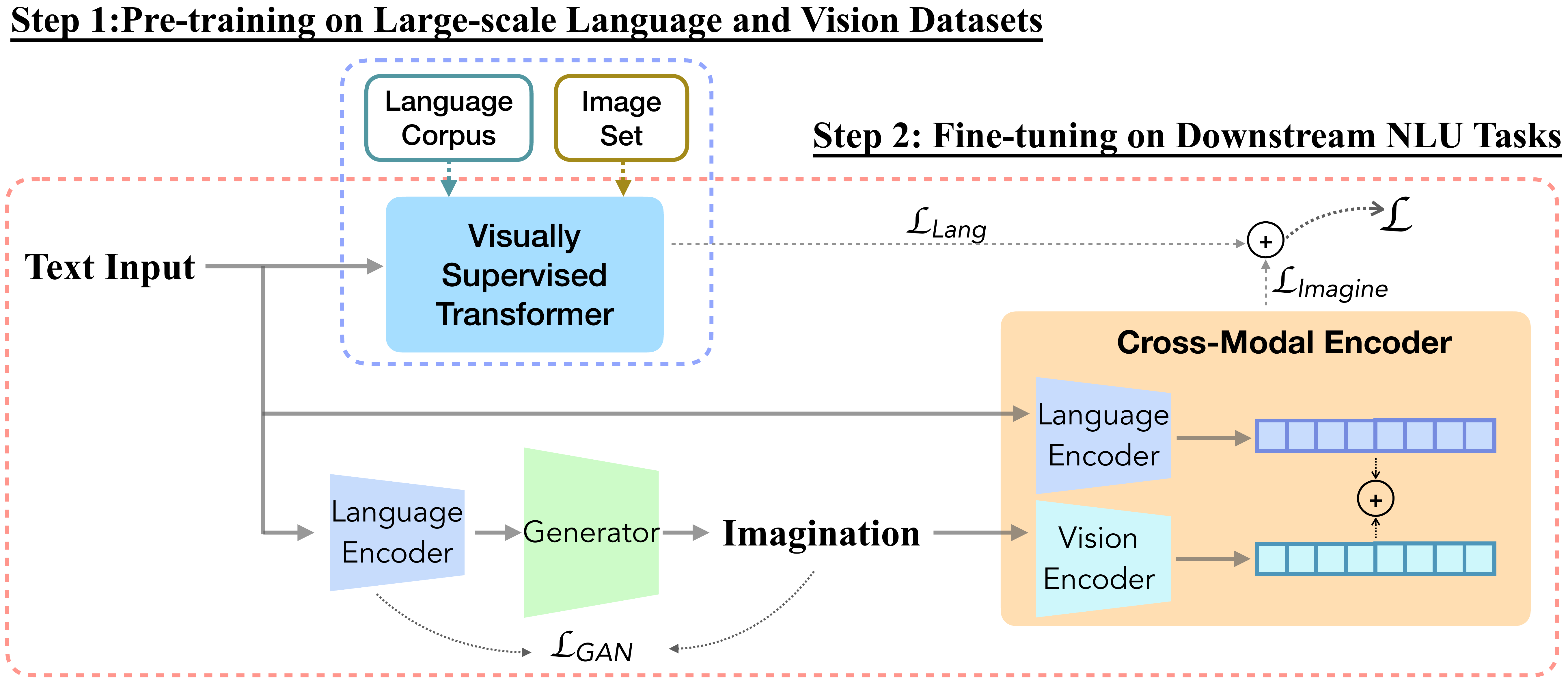}
    \caption{
    \textbf{Overview of \methodname~.}
    The generator $\mathit{G}$ visualize imaginations close to the encoded texts by minimizing $\mathcal{L}_{GAN}$.
    The cross-modal encoder $\mathit{E}_{c}$ learns imagination-augmented language representation.
    Two-step learning procedure consists of: 1) pre-train a Transformer with visual supervision from large-scale language corpus and image set, 2) fine-tune the visually supervised pre-trained Transformer and the imagination-augmented cross-modal encoder on downstream tasks.
    }
    \label{fig:overview}
\end{figure*}

We illustrate how we solve the existing text-only learning problem in natural language understanding tasks as the Imagination-augmented Cross-modal Language Understanding (\problemname~) problems in Section~\ref{sec:definition}.
Then we give a detailed illustration of our proposed \methodname~'s architecture in Section~\ref{sec:Model}.
Finally, we describe the procedure and training protocol of the perceptual-enhanced linguistic understanding paradigm in Section~\ref{sec:Procedure}.

\subsection{Problem Definition}
\label{sec:definition}
NLU is concerned with understanding the semantic meaning of the given utterances.
Data pieces for NLU can be structured as $(x_{context}, \gX, y)$, where $x_{context}$ represents the text context, $\gX =\{x_{1}, x_{2}, ..., x_{m}, m \in \mathbb{N}\}$ denote a set of text snippets, and $m$ denotes the number of text samples for a specific task.
The model learns to predict the ground truth label $y$, which is either regression or a classification label. \uu{While NLU is usually regarded as a language-only task, we attempt to solve it from a cross-modal perspective by introducing the novel \problemname~ problem.}

In our \problemname~ problem, data pieces are structured as $(x_{context}, i_{context}, \gX, \gI, y)$, in which $i_{context}$ represents the visual context related to the text context, and $\gI=\{i_{1}, i_{2}, ..., i_{n}, n \in \mathbb{N}\}$ denotes the imagination set. The ``imagination'' refers to the images that are visualized from the text. Here, $n$ is the number of visualized sentences for a specific task, which is the same as $m$ by default.

To solve this problem, we devise a novel \methodname~ to construct imagination from textual data and learn the bi-directional alignment between the imagination and text.
Specifically, for each piece of text $x_j$ in the sentence set $\gX$, we first follow \citep{esser2020taming,Radford2021LearningTV} and use a generative model to render a descriptive illustration $i_j$.
The visualized imagination will later serve as the visual input in the \problemname~ problem.

\subsection{Model Architecture}
\label{sec:Model}
\paragraph{Overview}
Figure~\ref{fig:overview} provides an overview of the \methodname~ framework.
\methodname~ consists of two modules: 1) the imagination generator $\mathit{G}$, 2) the imagination-augmented cross-modal encoder $\mathit{E}_{c}$.
Given the textual sentence $x=\{w_{1}, w_{2}, ..., w_{k}, k \in \mathbb{N}\}$ ($w_j$ denotes the $j$-th token in the sentence), $\mathit{G}$ generates corresponding visual imagination $i$.
The cross-modal encoder then encodes $x$ and $i$ as $\vt$ and $\vv$, respectively.
\methodname~ explicitly provides imagination supervision to the visually-supervised Transformer during fine-tuning on downstream NLU tasks.

\paragraph{Imagination Generator}
Previous studies introduce visual supervision through retrieval from the web or image sets. However, it is hard to find visuals that perfectly match the topics discussed in each text snippet, especially for the relatively complicated text input for the NLU tasks. Such misalignment between the input text and the retrieved visuals might hinder the model from general language understanding learning. Out of consideration for cross-modal feature alignment, we choose to render specific visualization corresponding to each piece of input text from scratch.
Specifically, we construct imagination of the textual input with a large-scale vision and language model guided generative framework -  VQGAN+CLIP~\citep{crowson2022vqgan}.
For each piece of input text $x$, we treat it as the prompt and use the VQGAN~\citep{esser2020taming} model to render the imagination $i$ with $128 \times 128$ resolution and $200$-step optimization. 
At each optimization step, we use the CLIP~\cite{Radford2021LearningTV} model to assess how well the generated image corresponds to the text.
\begin{equation}
\mathcal{L}_{GAN} = 2 [\arcsin(\frac{1}{2}\norm{\vt-\vv}) ] ^ 2
\label{eq:vqgan_clip_loss}
\end{equation}
To be specific, CLIP encodes the input text $x$ and the corresponding imagination $i$ as $\vt$ and $\vv$, and the training objective is to minimize the distance between $\vt$ and $\vv$ in the cross-modal embedding space.

\paragraph{Cross-modal Encoder}
We adopt CLIP as the cross-modal encoder to encode the input text and the generated imaginations. 
CLIP \cite{Radford2021LearningTV} is trained on large-scale image-text pairs and is able to align visual and textual input in the embedding space.
Specifically, we use the $ViT-B/32$ version of Vision Transformer as the image encoder, and Transformer \cite{Vaswani2017AttentionIA} with the architecture modifications described in \cite{Radford2019LanguageMA} as the text encoder.
For each modality, the self-attention (SA) module is applied to model the regions of imagination or the words of the text as follows:
\begin{equation}
SA(F) = concat(softmax \frac{FW_{j}^{Q}{FW_{j}^{K}}^{\mathrm{T}}}{\sqrt{d_{k}}}FW_{j}^{V}, ...) W
\label{eq:attention}
\end{equation}
where $F$ denotes the set of regions of the imagination or the words of the textual sentence. $W_{j}^{Q}$, $W_{j}^{K}$, and $W_{j}^{V}$ represents the weight in the $j$-th head for query, key and value respectively. $d_k$ is the dimension of the embedding. $W$ is the weight matrix for multiple heads.

To solve the \problemname~ problem, we learn the bi-directional relationship between the text input and the visualized imagination.
We apply late fusion on the text feature $\vt$ and visual feature $\vv$ to construct the cross-modal feature. Given the set of visual features $S_\vv$ and textual features $S_\vt$, the fused embedding $X_S$ can be given with:
\begin{equation}
X_S = [ReLU(W_t S_\vt + b_t), ReLU(W_j S_\vv + b_j)]
\label{eq:fusion}
\end{equation}
where $W$ and $b$ are  of  two separate fully connected layers to the visual and text embeddings.
The fused embeddings $X_S$ will go through two fully connected layers before we receive the final imagination-augmented language representation.

\paragraph{\upp{Visually-supervised Transformer}}
We implement the visually-supervised Transformer language model proposed in~\citet{Tan2020VokenizationIL}.
The model architecture is a BERT-like pure-language-based masked language model.

\subsection{Learning Procedure}
\label{sec:Procedure}
We introduce a novel paradigm to better understand natural language by incorporating existing language models with visual imagination.
As shown in Figure~\ref{fig:overview}, the procedure consists of two steps: (1) pre-train the visually-supervised Transformer, and (2) fine-tune the framework with imagination on downstream tasks.

\paragraph{Step 1: Visually-supervised Pre-training}
We pre-train a visually-supervised Transformer following the scheme proposed in VOKEN~\citep{Tan2020VokenizationIL}, which extrapolates cross-modal alignments to language-only data by contextually mapping language tokens to the related images.
In addition to masked language modeling, VOKEN proposed a voken classification task: given a set of tokens with masks, the model is asked to predict the best-matching image (the voken) for each tokens. The pre-training loss can be given as:
\begin{equation}
\mathcal{L} = - \lambda_1 \sum_{w_j \in \hat{s}}\log q_j(w_j|\check{s})- \lambda_2 \sum_{w_j \in \hat{s}}\log p_j(v(w_j;s)|\check{s})
\label{eq:pretrain_loss}
\end{equation}
Here $s$ is the token set, $\hat{s}$ is the masked tokens, and $\check{s}$ is the unmasked tokens.
The $q_j$ and $p_j$ represent the conditional probability distribution of the $j$-th token given the token $w_j$ and voken $v(w_j;s)$ respectively, and $\lambda_1$ and $\lambda_2$ are the balance factor of the masked language modeling task and the voken-classification task.
The cross-modal classification task enables the model to learn the matching between the tokens from the language corpus (e.g., wiki) and its most-related images from the image set (e.g., MSCOCO).

\paragraph{Step 2: Imagination-augmented Fine-tuning}
We use GLUE~\citep{Wang2018GLUEAM} and SWAG~\citep{Zellers2018SWAGAL} as the downstream datasets in the following sections.
Our proposed \methodname~ learns to minimize the cross-entropy loss below:
\begin{equation}
\mathcal{L}_{Imagine} = - \sum_{j=1}^{|D|} \sum_{k=1}^{K} y_k \log p_k (d_j(\vt;\vv) | D)
\label{eq:ima_loss}
\end{equation}
where $j$ denotes the $j$-th data sample in dataset $D$, and $K$ os the class number.
The $p_k$ represents the conditional probability distribution of $d_j$.
During fine-tuning, the visually-supervised Transformer language model only relied on the textual input to make predictions. The loss are computed as:
\begin{equation}
\mathcal{L}_{Lang} = - \sum_{j=1}^{|D|} \sum_{k=1}^{K} y_k \log p_k (d_j(\vt) | D)
\label{eq:lang_loss}
\end{equation}
Notice that we use MSE loss for the regression task.
The imagination-augmented loss and pure-language based loss are summed up with a balance factor $\lambda$ in a jointly training schema as:
\begin{equation}
\mathcal{L} = \lambda \mathcal{L}_{Imagine} + (1-\lambda) \mathcal{L}_{Lang}
\label{eq:total_loss}
\end{equation}
We use Adam Optimizer with a learning rate $1e-4$ for the GLUE benchmark and $2e-5$ for the SWAG dataset.
We discuss more details in Section~\ref{sec:experiments}.

%% file: sections/4-experiments.tex
\section{Experiments}
\label{sec:experiments}

\begin{table*}[h]
\centering
\resizebox{\textwidth}{!}{%
\begin{tabular}{l lll lll lll lll}
\toprule
&\multicolumn{3}{c}{\textbf{SST-2}} & \multicolumn{3}{c}{\textbf{QNLI}} &  \multicolumn{3}{c}{\textbf{QQP}}  &  \multicolumn{3}{c}{\textbf{MNLI}} \\
\cmidrule(lr){2-4}\cmidrule(lr){5-7}\cmidrule(lr){8-10}\cmidrule(lr){11-13}
\textbf{Extreme Few-shot} & $0.1\%$ & $0.3\%$ & $0.5\%$ & $0.1\%$ & $0.3\%$ & $0.5\%$ & $0.1\%$ & $0.3\%$ & $0.5\%$ & $0.1\%$ & $0.3\%$ & $0.5\%$ \\
    \midrule
    $VOKEN(Bert_{base})$      & 54.70 & 77.98 & 80.73 & 50.54 & 51.60 & 61.96 & 44.10 & 60.65 & 65.46 & 37.31 & 54.62 & 58.79 \\
    $\methodname~(Bert_{base})$       & \textbf{77.98} & \textbf{80.96} & \textbf{81.42} & \textbf{51.64} & \textbf{58.33} & \textbf{64.03} & \textbf{49.36} & \textbf{63.67} & \textbf{71.17} & \textbf{40.07} & \textbf{56.49} & \textbf{59.57} \\
    \hline
    $VOKEN(Roberta_{base})$   & 70.99 & 71.10 & 77.86 & 54.37 & 62.23 & 65.78 & 62.32 & 67.25 & 70.18 & 48.59 & 49.76 & 58.23 \\
    $\methodname~(Roberta_{base})$    & \textbf{75.34} & \textbf{78.66} & \textbf{83.60} & \textbf{54.79} & \textbf{65.03} & \textbf{65.83} & \textbf{65.43} & \textbf{68.11} & \textbf{70.77} & \textbf{48.94} & \textbf{52.74} & \textbf{59.39} \\
\midrule
\midrule
\textbf{Normal Few-shot} & $1\%$ & $3\%$ & $5\%$ & $1\%$ & $3\%$ & $5\%$ & $1\%$ & $3\%$ & $5\%$ & $1\%$ & $3\%$ & $5\%$ \\
    \midrule
    $VOKEN(Bert_{base})$      & 81.40 & 86.01 & 84.75 & 64.17 & 77.36 & 80.19 & 72.55 & 78.37 & 80.50 & 60.45 & 62.73 & 72.35 \\
    $\methodname~(Bert_{base})$       & \textbf{82.45} & \textbf{87.04} & \textbf{86.47} & \textbf{65.09} & \textbf{79.54} & \textbf{80.52} & \textbf{74.31} & \textbf{78.69} & \textbf{80.52} & \textbf{62.15} & \textbf{70.43} & \textbf{73.73} \\
    \hline
    $VOKEN(Roberta_{base})$   & 83.78 & 84.08 & 87.61 & 75.00 & 81.16 & 81.23 & 73.14 & 79.09 & 79.63 & 63.51 & 70.68 & 74.02 \\
    $\methodname~(Roberta_{base})$    & \textbf{83.83} & \textbf{84.63} & \textbf{89.11} & \textbf{79.35} & \textbf{81.41} & \textbf{81.65} & \textbf{73.72} & \textbf{79.38} & \textbf{79.81} & \textbf{65.66} & \textbf{70.76} & \textbf{74.10} \\
    \bottomrule
\end{tabular}
}
\caption{\textbf{Model-agnostic Improvement in Few-shot Setting.} \methodname~ and VOKEN upon BERT and RoBERTa base size architecture are fine-tuned in Extreme Few-shot ($0.1\%$, $0.3\%$, $0.5\%$) and Normal Few-shot setting ($1\%$, $3\%$, $5\%$). 
For the few-shot setting, we use large and stable datasets from GLUE Benchmark.
We compare accuracy on SST-2, QNLI, QQP, and MNLI and the average of accuracy and F1 score on QQP. \textbf{BEST} results are highlighted.
}
    \label{tab:fewshot}
\end{table*}

\subsection{Experimental Setup}

\paragraph{Datasets \& Metric}
We conduct experiments to evaluate the performance of our proposed method over SST-2~\citep{Socher2013RecursiveDM}, QNLI~\citep{rajpurkar-etal-2016-squad}, QQP~\citep{shanker2017qqp}, MultiNLI \cite{williams2018mnli}, MRPC~\citep{dolan2005automatically}, STS-B~\citep{eneko2007sts} from GLUE \cite{Wang2018GLUEAM} Benchmark, and SWAG \cite{Zellers2018SWAGAL} dataset.
We construct few-shot setting subsets by taking $0.1\%$, $0.3\%$, and $0.5\%$ of training instances as the Extreme Few-shot Setting, and  $1\%$, $3\%$, and $5\%$ as the Normal Few-shot Setting.
We train the model with the subsets and 
evaluate its performance on the complete development set. 
We use accuracy as the default evaluation metric and compare such results in the following sections.

\paragraph{Baselines}
\label{par:baselines}
We choose BERT \cite{Devlin2019BERTPO} and RoBERTa \cite{Liu2019RoBERTaAR} as the base language models, and apply our ~\methodname~ framework on top of their small and base architectures for comparison.
A recent study proposes a  visually-supervised language model VOKEN \cite{Tan2020VokenizationIL} that introduces visual supervision into language model pre-training by borrowing external knowledge from retrieved images of the tokens.
In natural language understanding tasks, VOKEN achieved improvements over language-based baselines BERT and RoBERTa.
Thus we also use VOKEN built upon these language-based models as a set of powerful baselines.
In the following experiments, each model is first pre-trained with visual supervision introduced in \cite{Tan2020VokenizationIL} upon the four base models (BERT$_{small}$, BERT$_{base}$, RoBERTa$_{small}$ and RoBERTa$_{base}$).
Then the models will be fine-tuned on downstream tasks.

Notice that base models and VOKEN use pure-language training objectives during fine-tuning.
Neither of them utilizes the visual signals inherent in the downstream language corpora.
In contrast, our \methodname~ explicitly introduces visual imagination supervisions into fine-tuning and inference stages.

\paragraph{Implementation Details}
We train RoBERTa with the same configurations as a robustly optimized pre-training approach based on BERT of the same size.
BERT$_{small}$ has $6$ repeating layers, $512$ hidden dimension.
BERT$_{base}$ has $12$ repeating layers, $768$ hidden dimension.

The imagination of the texts is generated interactively by using VQGAN+CLIP, with $128 \times 128$ size, $500$ iterations.
We use pre-trained VQGAN (imagenet$_{f16}$) and CLIP (ViT-B/$32$).
We leverage CLIP (ViT-B/$32$) as our language and vision model for premise and hypothesis, and imagination of them.
The text and image dimension is $512$.
The dropout rate is set to $0.1$. We use Cross-Entropy loss for our cross-modal classification.
Each model was first pre-trained on $4$ TITAN RX GPUs for $30$ epochs with early stopping and a batch size of $32$ and a sequence length of $126$.
The optimizer used is Adam with a learning rate of $2e-4$ and a weight decay of $0.01$.
The models are then fine-tuned on GLUE benchmark and SWAG dataset for $3$ epochs with $32$ batch size.
We adopt the joint training strategy for our proposed \methodname~ and visually supervised transformer during fine-tuning.
The learning rate of the Adam optimizer is set as $1e-4$ and $2e-5$ for GLUE and SWAG, respectively.

\begin{table*}[h]
\centering
\resizebox{\textwidth}{!}{%
\begin{tabular}{l l lll lll lll lll l}
\toprule
&& \multicolumn{3}{c}{\textbf{SST-2}} & \multicolumn{3}{c}{\textbf{QNLI}} &  \multicolumn{3}{c}{\textbf{QQP}}  &  \multicolumn{3}{c}{\textbf{MNLI}} & \textbf{ALL} \\
\cmidrule(lr){3-5}\cmidrule(lr){6-8}\cmidrule(lr){9-11}\cmidrule(lr){12-14}\cmidrule{15-15}
\textbf{Base Model}& \textbf{Method} & $0.1\%$ & $1.0\%$ & $3.0\%$ & $0.1\%$ & $1.0\%$ & $3.0\%$ & $0.1\%$ & $1.0\%$ & $3.0\%$ & $0.1\%$ & $1.0\%$ & $3.0\%$ & \textbf{Avg.}\\
    \midrule
    BERT$_{base}$ & Direction        & 49.01 & 79.59 & 87.15 & 51.31 & 52.55 & 66.90 & 56.74 & 31.58 & 31.59 & 32.73 & 61.54 & 70.72 & 55.95 \\
    BERT$_{base}$ & Unify            & 48.96 & 77.98 & 86.92 & 50.54 & 52.02 & 67.20 & 55.29 & 56.93 & 79.09 & 39.05 & 63.29 & 70.86 & 62.34 \\
    BERT$_{base}$ & \methodname~     & 77.98 & 82.45 & 87.04 & 51.64 & 65.09 & 79.54 & 49.36 & 74.31 & 78.69 & 40.07 & 62.15 & 70.43 & 68.23 \\
    \midrule
    RoBERTa$_{base}$ & Direction     & 72.71 & 80.38 & 84.63 & 54.91 & 74.68 & 78.58 & 61.57 & 74.68 & 31.59 & 32.95 & 61.96 & 70.62 & 64.94 \\
    RoBERTa$_{base}$ & Unify         & 75.11 & 80.04 & 88.07 & 53.62 & 74.64 & 78.47 & 64.94 & 74.85 & 76.84 & 51.12 & 65.42 & 70.74 & 71.15 \\
    RoBERTa$_{base}$ & \methodname~  & 75.34 & 83.83 & 84.63 & 54.79 & 79.35 & 81.41 & 65.43 & 73.72 & 79.38 & 48.94 & 65.66 & 70.76 & 71.93 \\
    \bottomrule
\end{tabular}
}
\caption{\textbf{Method Design Ablation in Few-shot Setting.} We compare the results of two variants over $0.1\%$, $1.0\%$, $3.0\%$ of SST-2, QNLI, QQP and MNLI dataset. Details of \textit{Direction} and \textit{Unify} are illustrated in Section~\ref{sec:ablation}.
}
    \label{tab:fewshot_ablation}
\end{table*}

\begin{table*}[h]
\centering
\resizebox{\linewidth}{!}{%
\begin{tabular}{l l llll llll l}
\toprule
& &\multicolumn{4}{c}{\textbf{Extreme Few-shot (0.1\%)}} & \multicolumn{4}{c}{\textbf{Normal Few-shot (3.0\%)}} & \textbf{ALL}\\
\cmidrule(lr){3-6}\cmidrule(lr){7-10}\cmidrule(lr){11-11}
\textbf{Base Model}&\textbf{Composition} & SST-2 & QNLI & QQP & MNLI & SST-2 & QNLI & QQP & MNLI & \textbf{Avg.}\\
    \midrule
    BERT$_{base}$& Textual-Only            & 49.08 &50.54 &55.48 &38.82 & 87.50 &67.05 &77.42 &71.00 &62.11\\
    BERT$_{base}$& Visual-Only             & 59.97 & 50.56 & 49.01 & 39.05 & 86.81 & 67.23 & 79.06 & 70.80 & 62.81\\
    BERT$_{base}$& Visual+Textual (VT)     & 53.89 & 50.54 & 49.15 & 38.83 & 87.04 & 66.81 & 79.16 & 70.77 & 62.02\\
    BERT$_{base}$& Bi-directional VT       & 77.98 & 51.64 & 49.36 & 40.07 & 87.04 & 79.54 & 78.69 & 70.43 & 66.84\\
    \hline
    RoBERTa$_{base}$& Textual-Only          & 75.57 &53.85 &64.96 &35.28 & 84.07 &78.51 &75.76 &51.48 &64.93\\
    RoBERTa$_{base}$& Visual-Only          & 75.11 & 54.18 & 65.01 & 47.22 & 84.17 & 79.88 & 76.88 & 70.56 & 69.12\\
    RoBERTa$_{base}$& Visual+Textual (VT)  & 74.20 & 53.98 & 65.43 & 47.35 & 83.94 & 79.96 & 76.87 & 70.73 & 69.05\\
    RoBERTa$_{base}$& Bi-directional VT    & 75.34 & 54.79 & 65.43 & 48.94 & 84.63 & 81.41 & 79.38 & 70.76 & 70.08\\
    \bottomrule
\end{tabular}
}
\caption{\textbf{Imagination Composition Ablation in Few-shot Setting.} \textit{Bi-directional VT} represents the full input for \methodname~. More details about \textit{Textual Only}, \textit{Visual Only} and \textit{Visual+Textual} are illustrated in Section~\ref{sec:ablation}.
}
    \label{tab:input_ablation}
\end{table*}

\subsection{Few-shot Learning Results}
We claim that introducing imagination into language processing helps the existing language-based system tackle the low-resource situation.
Thus, the automatically generated imagination helps reduce the human effort to annotate textual data.
To verify this, we define two situations, a normal few-shot setting, and an extreme few-shot setting.
For the normal few-shot setting, we keep $1\%$, $3\%$, and $5\%$ of the training dataset for each task in GLUE Benchmark.
For the extreme few-shot setting, we keep a lower number of the training dataset, which is reduced to $0.1\%$, $0.3\%$, and $0.5\%$ of the training dataset.
We train the models with the same configuration under these two settings and compare them with visually supervised transformer baselines to confirm the benefit that our proposed \methodname~ brings to the few-shot situation.

Results of the few-shot setting are reported in Table~\ref{tab:fewshot}.
Following \citet{Tan2020VokenizationIL}, we only report the four largest and stable tasks in GLUE for the model-agnostic comparison.
We report the accuracy for SST-2, QNLI, MNLI.
For QQP and MRPC, we report the average of F1 and accuracy.
For SWAG, we report the correlation.
We observe that the imagination information remarkably helps with both the normal few-shot curriculum and extreme few-shot curriculum.
\uuu{We assume the imagination-augmented fine-tuning successfully transfers the language understanding from the large-scale vision and language model.
Thus \methodname~ achieves consistent performance gain and shows great superiority of generalization and transferring ability.}

\begin{table*}[h]
\small
\centering
\begin{tabular}{l l l l l l l l l l l l}
\toprule
\textbf{Base Model} & \textbf{Method} & \textbf{SST-2} & \textbf{QNLI}  & \textbf{QQP} & \textbf{MNLI} & \textbf{MRPC} & \textbf{STS-B} & \textbf{SWAG} & \textbf{Avg.} \\
    \midrule
    BERT$_{small}$ & VOKEN         & 89.7 & 85.0 & 87.3 & 78.6 & 78.2 & 80.4 &  57.6 & 79.5\\
    BERT$_{small}$ & \methodname~   & \textbf{89.8} & \textbf{86.2} & \textbf{87.7} & \textbf{78.9} & \textbf{78.4} & \textbf{82.7} & \textbf{57.9} & \textbf{80.2}\\
    \hline
    BERT$_{base}$ & VOKEN        & \textbf{92.2} & 88.6 & 88.6 & 82.6 & 83.5 & 86.0 & 70.6 & 84.6\\
    BERT$_{base}$ & \methodname~  & 91.7 & \textbf{88.6} & \textbf{89.1} & \textbf{82.8} & \textbf{85.8} & \textbf{86.6} & \textbf{70.8} & \textbf{85.1} \\
    \midrule
    RoBERTa$_{small}$ & VOKEN         & 87.8 & 85.1 & 85.3 & 76.5 & 78.5 & 78.6 & 53.6 & 77.9\\
    RoBERTa$_{small}$ & \methodname~  & \textbf{89.2} & \textbf{85.1} & \textbf{86.5} & \textbf{76.8} & \textbf{79.0} & \textbf{78.7} & \textbf{53.7} & \textbf{78.3} \\
    \hline
    RoBERTa$_{base}$ & VOKEN          & 90.5 & \textbf{89.2} & 87.8 & 81.0 & 87.0 & 86.9 & 68.5 & 84.4\\
    RoBERTa$_{base}$ & \methodname~  & \textbf{91.6} & 89.1 & \textbf{87.9} & \textbf{82.6} & \textbf{87.7} & \textbf{86.9} & \textbf{68.5} & \textbf{84.9}\\
    \bottomrule
\end{tabular}
\caption{\textbf{Model-agnostic Improvement in Full Data Setting.} Results of \methodname~ and VOKEN upon BERT and RoBERTa of small($6L/512H$) and base($12L/768H$) architecture are reported. The models are fine-tuned over GLUE Benchmark and SWAG with access to the full dataset. \textbf{BEST} results are highlighted.
}
    \label{tab:modPer}
\end{table*}

\subsection{Ablation Studies}
\label{sec:ablation}
We conduct ablation studies over both the method side and data side to validate their contribution to our proposed \methodname~.
\paragraph{Method Design Ablation}
Two method variants of our imagination-augmented encoder are built as baselines to validate the importance of our bi-directional cross-modal imagination design in \methodname~.
The variants are built upon RoBERTa$_{base}$ and BERT$_{base}$ base models.
Specifically, we develop variant \textit{Direction} and \textit{Unify}.
\textit{Direction} represent alignment between text input and imagination into a directional embedding as FUSE($\vt_{sen1} - \vi_{sen1}$, $\vt_{sen2} - \vi_{sen2}$).
\textit{Unify} encode the text and imagination, considering the direction from vision to language by encoding as FUSE($\vt_{sent1}$, $\vt_{sent2}$, $\vi_{sent1}$, $\vi_{sent2}$).
While \textit{\methodname~} consider direction from visoin to language and language to vision by encoding as the combination of FUSE($\vt_{sent1}$, $\vi_{sent2}$) and FUSE($\vi_{sent1}$, $\vt_{sent2}$).
As shown in Table~\ref{tab:fewshot_ablation}, our bi-directional imagination and language learning achieve stable and best average performance.
These results indicate that our bi-directional imagination method design obtain generalization and transferring ability.
We assume \methodname~ benefits from both learning from language to vision and learning from vision to language simultaneously.

\paragraph{Imagination Composition Ablation}
The composition of the imagination is essential for the performance.
To further study the importance of full imagination, we ablate the data side by constructing a textual-only model denoted as \textit{Textual Only}, a visual-only imagination denoted as \textit{Visual Only} and a single directional imagination input denoted as \textit{Visual+Textual}.
\textit{Visual Only} and \textit{Visual+Textual} represent the imagination model use visual pairs $(\vi_{sent1}, \vi_{sent2})$ and one direction visual and textual pairs $(\vi_{sent1}, \vt_{sent2})$ as input respectively. Our full approach use \textit{Bi-directional VT} which takes $(\vi_{sent1}, \vt_{sent2})$ and $(\vt_{sent1}, \vi_{sent2})$ as input.

Results are reported in Table~\ref{tab:input_ablation} for Extreme Few-shot setting and  normal few-shot setting.
We observe \textit{Bi-directional VT} data input achieve the most stable and the best average performance.
Results show the importance of bi-directional imagination from all the textual input to construct an imagination-augmented cross-modal encoder.

\begin{figure*}[t]
    \centering
    \includegraphics[width=\linewidth]{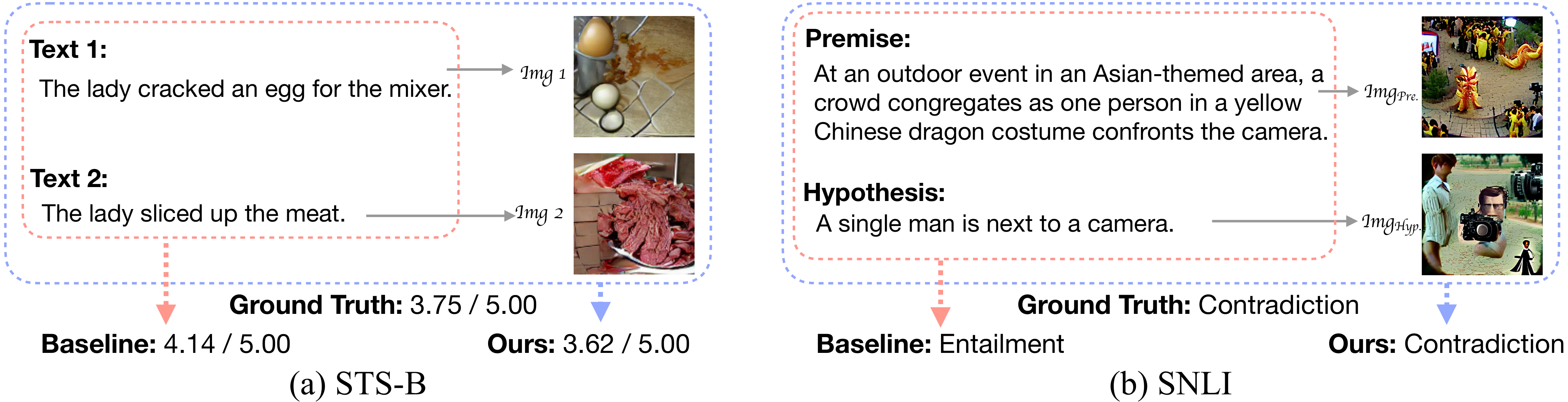}
    \caption{Case studies on the STS-B and SNLI tasks. The baseline models yield predictions solely based on the text input, while our approach takes both the text input and corresponding visualization into consideration. On both tasks, our ~\methodname~ gives predictions that are more aligned with the ground truth.
    }
    \label{fig:case_studies}
\end{figure*}

\subsection{Model-agnostic Improvement}
\methodname~ is a model-agnostic training paradigm that could help existing models achieve consistent gain over GLUE and SWAG with both the few-shot setting and full data setting.
To validate such model-agnostic effectiveness of our proposed novel paradigm in processing natural language, we compare the performance with two language models (BERT and RoBERTa) of two architectures ("6L/512H" and "12L/768H"), and a strong visually supervised pre-trained baseline VOKEN \cite{Tan2020VokenizationIL}.

Table~\ref{tab:modPer} shows the metric comparison on GLUE and SWAG.
The base models are trained with a masked language model.
The VOKEN model is pre-trained with a masked language model with an additional voken-classification task as introduced visual supervision.
\methodname~ achieves model-agnostic improvement over the model that solely fine-tune based on textual information, including the pure-language-based model and visually supervised pre-trained model.
The gain is consistently observed from different architectures of models.

%% file: sections/5-analysis.tex
\subsection{Case Study}
Figure~\ref{fig:case_studies} lists out our examples for the case study.
We show the results from the natural language inference and sentence similarity task.
We use examples from the STS-B and SNLI datasets.
Our contextual imagination describes the textual input as expected and provides an external prediction reference.

For example (a), given the structurally diversified sentence and low $n$-grams overlaps but high semantic similarity, we observe that the pure language-based model predicts the wrong label.
While the imagination helps the model capture the semantic similarity between two textual inputs via comparing the cross-modal semantics with the imagination information.
From example (b), we observe the pure language-based model predicts the wrong label based on the similar sentence structure and high $n$-grams overlaps.
While the imagination helps the model capture the difference between the similar premise and hypothesis text.

%% file: sections/6-conclusion.tex
\section{Conclusion}
We treat the text-only learning problem in Natural Language Understanding tasks as a cross-modal language understanding problem with generated imagination as supervision.
In this scenario, the task aims to bridge the gap between the human and the agent language understanding in both linguistic and perceptual procedures.
To address the proposed problem, we devised a model-agnostic learning paradigm \methodname~.
Specifically, we build the imagination of the downstream dataset using an interactive generative approach with guidance from a self-supervised pre-trained large-scale image and text model.
Our proposed \methodname~ surpassed baselines of two architecture sizes by a large margin in the few-shot setting.
The improvement is consistently observed over pure-language baselines (BERT and RoBERTa) and visually supervised VOKEN on the GLUE and SWAG dataset.
The results show the superiority of our \methodname~ in language understanding and handling low-resource circumstances.

\section*{Acknowledgement}
We would like to thank the Robert N. Noyce Trust for their generous gift to the University of California via the Noyce Initiative.
The research was also sponsored by the U.S. Army Research Office and was accomplished under Contract Number W911NF19-D-0001 for the Institute for Collaborative Biotechnologies.
The writers' opinions and conclusions in this publication are their own and should not be construed as representing the sponsors or the U.S. government's official policy, expressed or inferred. Regardless of any copyright notation herein, the United States government is authorized to reproduce and distribute reprints for government purposes.

\section*{Ethical Statement}
In this study, we only cover NLU datasets with English annotations. Such limitation is since the large-scale pre-trained multimodal models used in our studies, such as CLIP and VQGAN, are only trained on English corpus as of the date we conduct the experiments~\footnote{As of Dec. 2021.}.

This study use CLIP and VQGAN to render images given the text prompt. 
Suppose there exists any bias in the training dataset for the large-scale pre-trained multimodal models used in our study. In that case, our ``imagination'' approach may face an issue of fairness since the visual generative model might be more likely to illustrate specific types of images that it has seen in the training data. 
Moreover, if the training dataset for CLIP or VQGAN contains any personal information, then our ``imagination'' approach may strike a threat on privacy leakage given certain triggers or prompts. 
Even though we did not witness such issues in our study, we should keep in mind that the aforementioned behaviors would impair \methodname~'s effectiveness.